\documentclass[10pt,twocolumn,letterpaper]{article}

\usepackage[pagenumbers]{cvpr} 

\usepackage[dvipsnames]{xcolor}
\usepackage{makecell}
\usepackage{booktabs}

\usepackage{times}
\usepackage{epsfig}
\usepackage{graphicx}
\usepackage{amsmath}
\usepackage{amssymb}
\usepackage{booktabs}
\usepackage[utf8]{inputenc} 
\usepackage[T1]{fontenc}    
\usepackage{amsfonts}       
\usepackage{nicefrac}       
\usepackage{microtype}      
\usepackage{xcolor}         

\usepackage{tabularray}
\UseTblrLibrary{booktabs}

\usepackage{times}
\usepackage{epsfig}
\usepackage{commath}
\usepackage{wrapfig}
\usepackage{mathtools,amsbsy}
\usepackage{amsthm}
\usepackage{wrapfig}

\usepackage{algorithm}
\usepackage{algpseudocode}

\usepackage{lipsum}

\DeclareMathOperator{\SO}{\textup{SO}}
\DeclareMathOperator{\Og}{\textup{O}}

\DeclarePairedDelimiterX{\myfloor}[1]{\lfloor}{\rfloor}{#1}

\usepackage{amsmath}

\definecolor{cvprblue}{rgb}{0.21,0.49,0.74}
\usepackage[pagebackref,breaklinks,colorlinks,citecolor=cvprblue]{hyperref}

\def\paperID{5127} 
\def\confName{CVPR}
\def\confYear{2024}

\title{TetraSphere: A Neural Descriptor for O(3)-Invariant Point Cloud Analysis}

\author{Pavlo Melnyk, \quad Andreas Robinson, \quad Michael Felsberg, \quad Mårten Wadenbäck\\
{\small Computer Vision Laboratory, Department of Electrical Engineering, Linköping University, Sweden}\\
{\small \texttt{\{pavlo.melnyk, andreas.robinson, michael.felsberg, marten.wadenback\}@liu.se}}
}

\begin{document}
\maketitle
\begin{abstract}
In many practical applications, 3D point cloud analysis requires rotation invariance.
In this paper, we present a learnable descriptor invariant under 3D rotations and reflections, \ie, the $\Og(3)$ actions, utilizing the recently introduced steerable 3D spherical neurons and vector neurons. 
Specifically, we propose an embedding of the 3D spherical neurons into 4D vector neurons, which leverages end-to-end training of the model.
In our approach, we perform TetraTransform---an equivariant embedding of the 3D input into 4D, constructed from the steerable neurons---and extract deeper $\Og(3)$-equivariant features using vector neurons.
This integration of the TetraTransform into the VN-DGCNN framework, termed \textbf{TetraSphere}, negligibly increases the number of parameters by less than 0.0002\%.
TetraSphere sets a new state-of-the-art performance classifying randomly rotated real-world object scans of the challenging subsets of ScanObjectNN.
Additionally, TetraSphere outperforms all equivariant methods on randomly rotated synthetic data: classifying objects from ModelNet40 and segmenting parts of the ShapeNet shapes.
Thus, our results reveal the practical value of steerable 3D spherical neurons for learning in 3D Euclidean space.
The code is available at \url{https://github.com/pavlo-melnyk/tetrasphere}.
\end{abstract}    
\vspace{-8pt}
\section{Introduction}
\label{sec:intro}
\vspace{-3pt}
Automatic processing of 3D data obtained with sensors such as LIDARs, sparse stereo, and sparse time-of-flight is a central problem for many autonomous systems \cite{Hu_2022_CVPR, Fazlali_2022_CVPR, Sautier_2022_CVPR}.
Point clouds---in the form of an array of a fixed number of 3D coordinates and corresponding optional features (\eg, color or intensity)---are a common representation of such data in various 3D vision tasks.

Consider, for example, the task of 3D object classification, where the goal is to predict the correct class given a point cloud.
Importantly, the order of the points and different orientations of the shape do not alter its class membership. 
This imposes the requirements of permutation and rotation invariance on the classifier.
Furthermore, in certain real-world scenarios (such as left- and right-hand traffic), global reflection invariance is desired. 
For instance, a vehicle designed for either type of traffic may be considered the same.

Fulfilling the first requirement is commonly done by constructing a model using shared multilayer perceptrons (MLPs) and a global aggregation function, producing permutation-invariant features, as in, \eg, PointNet \cite{qi2017pointnet}. 

To attain rotation invariance \cite{van1995vision}, a common approach is to augment available data by performing random rotations and train the model in the hope that it can generalize to other, possibly unknown, orientations during inference.
However, such an approach relies heavily on augmentation and requires an increased model capacity. 
Such methods are commonly referred to as rotation-sensitive, \eg, \cite{qi2017pointnet++, wang2019dynamic}.
Using the augmentation approach with rotation-sensitive methods only \textit{approximates} rotation invariance.
\begin{figure}
	\centering
	\includegraphics[width=1\linewidth]{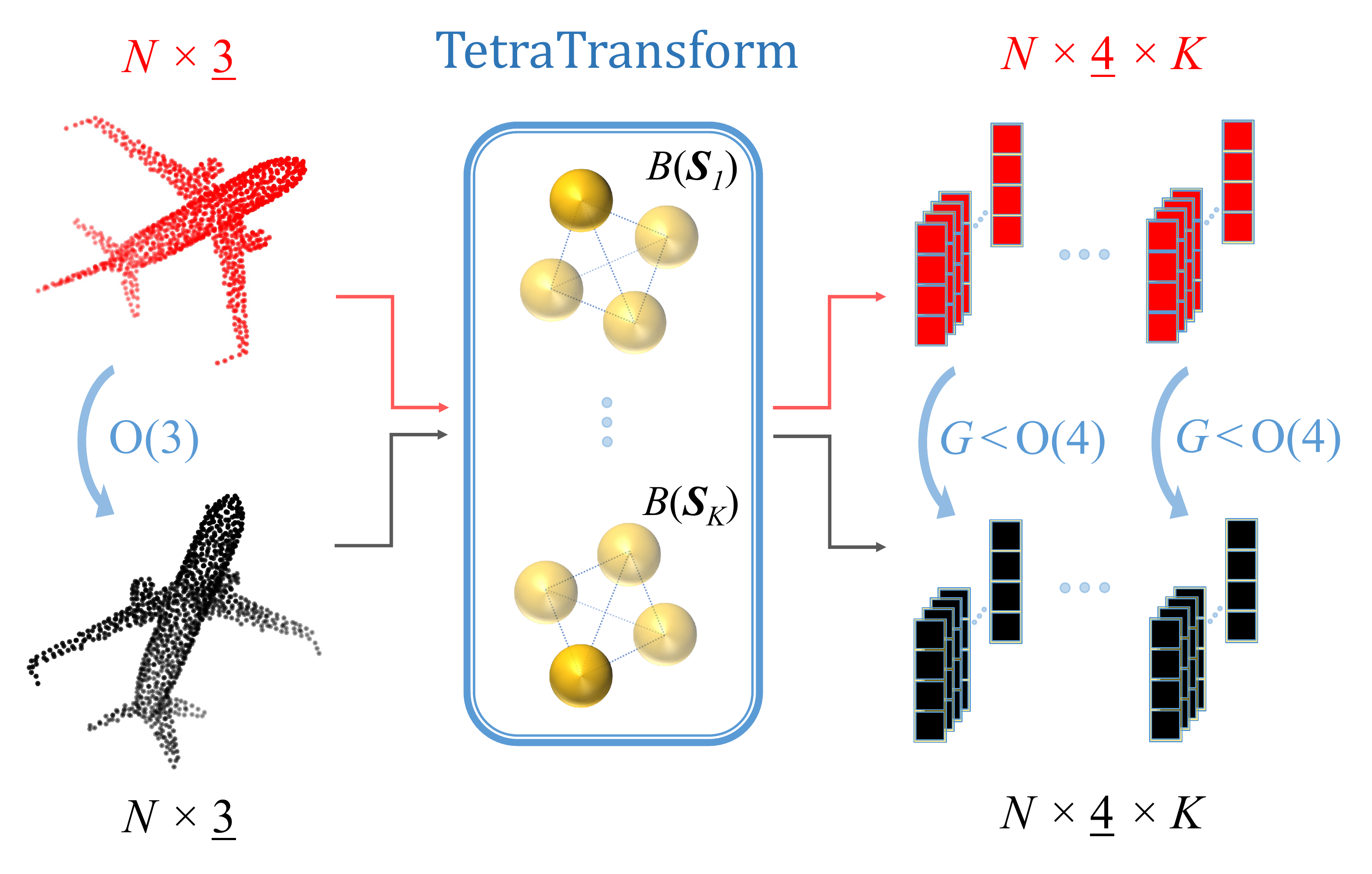}
	\caption{Key component in our method (best viewed in color): a learnable O(3)-equivariant TetraTransform layer consisting of $K$ steerable 3D spherical neurons \cite{melnyk2022steerable} that lifts the input 3D points to equivariant 4D representations
	 (see Section~\ref{sec:so3-equiv_features} for details).}
  \vspace{-12pt}
	\label{fig:tetratransform}
\end{figure}
There are also rotation-equivariant methods \cite{weiler20183d, thomas2018tensor, deng2021vector}, in which the learned features rotate correspondingly with the input, and rotation-invariant (RI) techniques \cite{xu2021sgmnet, li2021rotation, chen2022devil, chou20213dgfe, gu2022elganet}, in which the central trend is to construct RI low-level geometric features and use them instead of point coordinates.
An alternative approach is to compute a canonical pose and then de-rotate the input point cloud and perform processing on it \cite{fang2020rotpredictor,sun2021canonical,li2021closer}.

Our method is a combination of SO(3)-equivariant steerable 3D spherical neurons \cite{melnyk2022steerable} and vector neurons\cite{deng2021vector}, where deep rotation-equivariant features are learned using vector neurons and invariant predictions are obtained by taking the inner products of these features \textit{point-wise}.
However, unlike the original SO(3)-equivariant framework \cite{deng2021vector}, we propagate equivariant features through the network by constructing a specific 4D space spanned by what we call a \textit{tetra-basis}, as shown in Figure~\ref{fig:tetratransform}.
Our main hypothesis is that features from learned rotation-equivariant \textit{TetraTransform} projections are more expressive than the points themselves.

We summarize our contributions as follows:\newline
\noindent\textbf{(1)} We propose an embedding of 3D spherical neurons \cite{melnyk2022steerable} into 4D vector neurons \cite{deng2021vector}, which we show they are both O(3)-equivariant, and propose TetraSphere---a learnable O(3)-invariant descriptor for 3D point cloud classification, built upon VN-DGCNN \cite{deng2021vector}. 

\noindent\textbf{(2)} We unveil the practical utility of the steerable neurons, which, to the best of our knowledge, have never been used in an end-to-end framework previously.

\noindent\textbf{(3)} We demonstrate the effectiveness of TetraSphere by evaluating it on standard benchmarks, consistently outperforming the baseline VN-DGCNN, and setting new state-of-the-art performance classifying arbitrarily rotated real-world scans from ScanObjectNN~\cite{uy-scanobjectnn-iccv19}, even when they are significantly perturbed and occluded, and the best performance among equivariant methods benchmarked with the randomly rotated synthetic data from ModelNet40~\cite{wu20153d} and ShapeNet~\cite{chang2015shapenet}.

\section{Related Work}
\label{sec:related_work}
\subsection{Rotation-sensitive 3D point cloud learning}
PointNet \cite{qi2017pointnet} is the pioneering work for learning on raw point sets as input data for the tasks of classification, part segmentation, and semantic segmentation. 
Its limited ability for recognizing fine-grained patterns was addressed in the PointNet++ method \cite{qi2017pointnet++} that recursively applies PointNet on a nested partitioning of the input point cloud. 
Other noteworthy methods include PointCNN \cite{li2018pointcnn} with a special type of convolution operator applied to the input points and features before they are processed by an ordinary convolution, and dynamic graph CNN (DGCNN) \cite{wang2019dynamic}, where a graph convolution is applied to edges of the \textit{k}-nearest neighbor graph of the point clouds.
Xiang \emph{et~al.\@} \cite{Xiang_2021_ICCV} introduced CurveNet based on a sequence-of-points (curve) grouping operator and a curve aggregation operator. A more geometrically inspired approach was presented by Melnyk \emph{et~al.\@} \cite{melnyk2020embed}, who revisited modeling spherical decision surfaces with conformal embedding \cite{perwass2003spherical} in the context of learning 3D point cloud representations.

Somewhat surprisingly, similar to the projective method for 3D semantic segmentation by Järemo Lawin \emph{et~al.\@}~\cite{lawin2017deep}, it was shown by Goyal \emph{et~al.\@} \cite{goyal2021revisiting} that on a point cloud classification task, a simple projection-based baseline called SimpleView performs on par with 3D approaches. 
Moreover, the authors designed a protocol for a fair comparison between point cloud learning methods revealing the importance of many factors independent of the proposed architectures, such as evaluation procedure and hyperparameter tuning. 
Recently, a transformer-based approach combining local and global attention mechanisms was presented by Berg \textit{et al.} \cite{berg2022points}.

Notably, the aforementioned approaches are rotation-variant, \ie, they require data augmentation if rotation invariance is desired. 
This also entails the model having an increased number of parameters for memorizing the data in various orientations.
\subsection{Rotation-aware models}
\vspace{-5pt}
As an alternative, approaches have been proposed for learning rotation equivariant features, in which learned representations rotate in accordance with the input \cite{thomas2018tensor, fuchs2020se3, bokman2022zz, zhao2020quaternion, poulenard2021functional, luo2022equivariant}.
Among these are quaternion-based models \cite{zhao2020quaternion, shen20203d}
and methods that perform a projection of the 3D input to a unit sphere \cite{cohen2018spherical,Esteves_2018_ECCV} and realize convolutions in the spherical harmonic domain. 

The work of Deng \emph{et~al.\@}~\cite{deng2021vector} introduced \textit{vector neurons} by extending neurons from 1D scalars to 3D vectors, and thereby enabling a simple mapping of SO(3)-actions to latent spaces in the general rotation-equivariant framework.
In the context of equivariant methods, Melnyk \textit{et al.} proposed steerable 3D spherical neurons \cite{melnyk2022steerable}, which are SO(3)-equivariant filter banks obtained by virtue of conformal modeling \cite{perwass2003spherical, melnyk2020embed} and the symmetries of spheres as geometric entities \cite{melnyk2022steerable}.

Other methods make use of group representation theory and transform the input points into a space in which it is easier to express rotation-equivariant maps \cite{ thomas2018tensor, fuchs2020se3, poulenard2021functional}, and after that obtain rotation-invariant prediction, \eg, when performing classification. 
This is achieved using filters constrained to be combinations of spherical harmonics, which limits their expressiveness.
Therefore, such methods have naturally limited learning capability, and their performance falls short compared to rotation-sensitive methods for tasks that do not require rotation invariance.

There is a plethora of conceptually different works on hand-crafting low-level rotation invariant (RI) geometric features for arbitrary pairs of points (PPF) based on angles and distances \cite{zhang2020learning, zhao2019rotation, xu2021sgmnet, chou20213dgfe, gu2021lganet, gu2022elganet}, proposed to be used instead of the input point coordinates. 
For instance, similar to the triplets used by Granlund \emph{et~al.\@} \cite{granlund2004unrestricted}, Zhang \emph{et~al.\@} \cite{zhang2019rotation} introduced a convolution operator that uses a point neighborhood constructed with triple-point (reference-neighbor-centroid) local triangles. In contrast, vector norm and relative angles between points were used by Chen \emph{et~al.\@} \cite{chen2019clusternet}.
A robust RI representation, capturing both local and global shape structures, and region relation convolution, alleviating global information loss, were presented by Li \emph{et~al.\@} \cite{li2021rotation}.

The pose information loss problem was revealed and addressed by introducing a pose-aware RI convolution (PaRI-Conv) with compact and efficient kernels by Chen and Cong~\cite{Chen_2022_CVPR}. 
Therein, a lightweight augmented PPF (APPF) is proposed, encoding the local pose of each point in a local neighborhood in an ambiguity-free manner.
Notably, their approach is also invariant under reflections, \ie, O(3)-invariant, and they use local reference frames (LRFs) as input. 
However, utilizing principal component analysis (PCA) to construct the LRF for RI point cloud learning, as done by Kim \emph{et~al.\@}~\cite{kim2020rotation} and Xiao \emph{et~al.\@}~\cite{9102947}, is sensitive to perturbations. 
This is why Chen and Cong~\cite{Chen_2022_CVPR} proposed to build the LRFs upon local geometry only.

Input canonicalization is another category of methods that includes both rotation-variant (\eg, variants of \cite{qi2017pointnet, wang2019dgcnn} that use spatial transformers), -equivariant (\eg, \cite{fang2020rotpredictor, sun2021canonical, spezialetti2020learning}), and -invariant \cite{li2021closer} methods. 
The key idea in these approaches is to bring the input to a computed or predicted canonical reference frame and process it there.

Recently, Yu~\etal \cite{yu2023rethinking} utilized the point-cloud registration approach to achieve rotation invariance. They proposed registering the deep features to rotation-invariant features at intermediate levels in their Aligned Integration Transformer (AIT), thereby increasing feature similarities in the embedding space and attaining rotation invariance.

Our approach builds upon the equivariant framework \cite{deng2021vector}: we apply steerable 3D spherical neurons \cite{melnyk2022steerable} to learn in $K$ different spaces O(3)-equivariant 4D features from the 3D input point coordinates, and then aggregate the result by means of an equivariant pooling over $K$; finally, we propagate through the VN-backbone, wherein the inner product of these features in the equivariant feature space is computed (see Figure~\ref{fig:tetrasphere-b}).
This way, we create a learnable O(3)-invariant descriptor, encoding both unambiguous pose information and local and global context.

\section{Preliminaries}
\label{sec:background}
\vspace{-5pt}
In this section, we introduce the necessary notation and recap the notion of equivariance and invariance and theoretical results from prior work, which will enable us to realize an embedding of steerable 3D spherical neurons into 4D vector neurons.

We define a 3D point cloud $\mathcal{X} \in \mathbb{R}^{N \times (3+C)}$ as a collection of $N$ points, represented by their coordinates $\textbf{\textup{x}} \in \mathbb{R}^3$ concatenated with the corresponding optional features $\textbf{\textup{q}} \in \mathbb{R}^C$: $\mathcal{X} = \left\{{\textbf{\textup{x}}_n \oplus \textbf{\textup{q}}_n }\right\}_{n=1}^{N}$. 
In the scope of this paper, we focus only on the point coordinates and assume that the optional features are rotation- and reflection-invariant.

\subsection{Equivariance and invariance}
\label{sec:equivariance}
Given a group $G$ and a set of transformations $T_g : \mathcal{X} \rightarrow \mathcal{X}$ for $g \in G$, a function $f : \mathcal{X} \rightarrow \mathcal{Y}$ is said to be  $G$-\textit{equivariant} if for every $g$, there exists a transformation $V_g : \mathcal{Y} \rightarrow \mathcal{Y}$ such that 
\begin{equation}
\label{eq:equivariance}
    V_g[f(\textbf{\textup{x}})] = f(T_g[\textbf{\textup{x}}]) \text{\quad for all~} g \in G, \; \textbf{\textup{x}} \in \mathcal{X},
\end{equation}
where $T_g$ represents transformation parameters. 

Invariance is a particular type of equivariance. 
A function $f : \mathcal{X} \rightarrow \mathcal{Y}$ is said to be  $G$-\textit{invariant} if for every $g \in G$, the transformation $V_g : \mathcal{Y} \rightarrow \mathcal{Y}$ is the identity, \ie,
\begin{equation}
\label{eq:invariance}
    f(\textbf{\textup{x}}) = f(T_g[\textbf{\textup{x}}]) \text{\quad for all~} g \in G, \; \textbf{\textup{x}} \in \mathcal{X}.
\end{equation}

In particular, we consider invariance under 3D orthogonal transformations (rotations and reflections), \ie, the group O(3), and, as an intermediate step, equivariance under 3D rotations---the group $\SO$(3). 
In order to act as a transformation $T_g$ on a 3D vector $\textbf{\textup{x}} \in \mathbb{R}^3$, the elements $g \in \SO$(3) are often represented by $3\times 3$ rotation matrices $\textbf{\textit{R}}$ \cite{chirikjian2000engineering}. 
However, this representation is not unique \cite{zhou2019continuity}.

Our proposed descriptor, which we present in Section~\ref{sec:methodology}, is O(3)-invariant and equivariant under permutations of the input points.
That is, permuting point indices $1, \dots, N$ results in the corresponding permutation of the descriptor outputs. 

In the remainder of the manuscript, we use the same notation to represent a 3D rotation matrix $\textbf{\textit{R}}$ in the Euclidean space $\mathbb{R}^3$, the projective (homogeneous) space $\textit{P}(\mathbb{R}^3) \subset \mathbb{R}^4$, and $\mathbb{R}^5$, by appending the required number of ones to the diagonal of the original rotation matrix without changing the transformation itself \cite{melnyk2020embed}.

\subsection{Spherical neurons}
\label{sec:spherical_neurons}
\textit{Spherical neurons} are defined as neurons with (hyper)spherical decision surfaces \cite{perwass2003spherical, melnyk2020embed}.
Following Perwass \emph{et~al.\@} \cite{perwass2003spherical}, one embeds both a data vector $\textbf{x} \in\mathbb{R}^n$ and a hypersphere $(\textbf{c}, r)$ in $\mathbb{R}^{n+2}$ as
\begin{equation}
	\label{hypersphere_in_r}
	\begin{aligned}
		\textbf{\textit{X}} &= \big(x_1, \dots, x_n, -1, -\frac{1}{2}\lVert\textbf{x}\rVert^2\big)\in\mathbb{R}^{n+2},\\
		\textbf{\textit{S}} &= \big(c_1, \dots, c_n, \frac{1}{2}(\lVert\textbf{c}\rVert^2 - r^2), 1\big)\in\mathbb{R}^{n+2},
	\end{aligned}
\end{equation}
where  $\textbf{c} =(c_1, \dots, c_n)\in\mathbb{R}^n$ is the hypersphere center and $r\in\mathbb{R}$ is its radius.
Their scalar product in $\mathbb{R}^{n+2}$ is given by
\begin{equation}
    \label{scalar_product_isomorphism}
    \textbf{\textit{X}}^\top \textbf{\textit{S}} = -\frac{1}{2}\norm{\textbf{x}-\textbf{c}}^2 + \frac{1}{2}r^2~.
\end{equation} 
The sign of this scalar product depends on the relative position of the point to the sphere in the Euclidean space $\mathbb{R}^{n}$: inside the sphere if positive, outside of the sphere if negative, and on the sphere if zero \cite{perwass2003spherical}. 
Perwass \emph{et~al.\@} \cite{perwass2003spherical} suggested to use the scalar product \eqref{scalar_product_isomorphism} as a classifier, \ie, a \textit{spherical neuron} $f_{S}(\textbf{\textit{X}}; \textbf{\textit{S}}) = \textbf{\textit{X}}^\top \textbf{\textit{S}}$ with learnable parameters $\textbf{\textit{S}}\in \mathbb{R}^{n+2}$.
 Importantly, as noted by Melnyk \emph{et~al.\@}~\cite{melnyk2020embed}, spherical neurons do not necessarily require an activation function, due to the non-linearity of the embedding \eqref{hypersphere_in_r}.
 
During training, the components of $\textbf{\textit{S}}$ in \eqref{hypersphere_in_r} are treated as independent learnable parameters. 
Therefore, a spherical neuron effectively learns \textit{non-normalized} hyperspheres of the form $\widetilde{\textbf{\textit{S}}} = ({s_1}, \dots, {s_{n+2}}) \in \mathbb{R}^{n+2}$.
 Due to the chosen representation \cite{perwass2003spherical}, both normalized and non-normalized hyperspheres represent the same decision surface, and the spherical neuron can thus be written as
 \begin{equation}
    \label{eq:spherical_neuron}
    f_{S}\,(\textbf{\textit{X}}; \widetilde{\textbf{\textit{S}}}) = \textbf{\textit{X}}^\top \widetilde{\textbf{\textit{S}}} =  \gamma\, \textbf{\textit{X}}^\top \textbf{\textit{S}},
 \end{equation}
 where $\gamma:={s_{n+2}}$ is the (learned) normalization parameter and $\textbf{\textit{S}}\in\mathbb{R}^{n+2}$ is the normalized sphere defined in \eqref{hypersphere_in_r}.
From this point, we will write $\textbf{\textit{S}}$ when referring to a spherical decision surface, specifying its normalization if needed.

 Further details are found in the work of Melnyk \emph{et~al.\@} \cite{melnyk2020embed}, where, inter alia, it is demonstrated that the spherical neuron activations are isometries in 3D.
That is, rigid transformations commute with the application of the spherical neuron.
This result is a necessary condition to design rotation equivariant feature extractors based on spherical neurons \cite{melnyk2022steerable}, that we review in Section~\ref{sec:steerable_3d_neurons}. 

\subsection{Steerable 3D spherical neurons}
\label{sec:steerable_3d_neurons}
A steerable 3D spherical neuron, recently introduced by Melnyk \emph{et~al.\@} \cite{melnyk2022steerable}, is a filter bank consisting of
one learnable spherical decision surface $\textbf{\textit{S}} \in \mathbb{R}^5$ \eqref{hypersphere_in_r} and three copies: The original (learned) sphere center $\textbf{c}_0$ is first rotated to $\frac{\norm{\textbf{c}_0}}{\sqrt{3}}\,(1,1,1)$ with the corresponding (geodesic) rotation denoted as $\textbf{\textit{R}}_O$. 
The resulting sphere is then rotated into the other three vertices of the regular tetrahedron.
This is followed by rotating all four spheres back to the original coordinate system.
One steerable 3D spherical neuron is thus composed as the $4 \times 5$ matrix
\begin{equation}
	\label{eq:sphere_filter_bank}
	B(\textbf{\textit{S}}) = 
	\begin{bmatrix}
		(\textbf{\textit{R}}_O^{\top}\, \textbf{\textit{R}}_{T_i}\, \textbf{\textit{R}}_O\, \textbf{\textit{S}})^{\top} \\
	\end{bmatrix}_{i={0\ldots3}} ~,
\end{equation}
where each of $\{\textbf{\textit{R}}_{T_i}\}_{i=0}^{3}$ is the isomorphism in $\mathbb{R}^5$ corresponding to a 3D rotation from $(1,1,1)$ to the vertex $i+1$ of the regular tetrahedron. Hence, $\textbf{\textit{R}}_{T_0}=\textbf{I}_5$, \ie, $\textbf{\textit{S}}$ remains at $\textbf{c}_0$.

We can view the steerable spherical neuron \eqref{eq:sphere_filter_bank} as a function $f_{4S}(\,\cdot\,;\textbf{\textit{S}}): \mathbb{R}^5 \rightarrow \mathbb{R}^4$ with five learnable parameters as a vector $\textbf{\textit{S}}$. 
Crucially for our work, Melnyk \emph{et~al.\@}~\cite{melnyk2022steerable} proved that it is equivariant under 3D rotations:
\begin{equation}
\label{eq:filter_bank_equivariance}
    V_{\textbf{\textit{R}}} \, B(\textbf{\textit{S}}) \, \textbf{\textit{X}} = B(\textbf{\textit{S}})\,\textbf{\textit{R}}\textit{\textbf{X}},
\end{equation}
where ${\textbf{\textit{X}}} \in \mathbb{R}^{5}$ is a properly embedded 3D input point, $\textbf{\textit{R}}$ is a representation of the 3D rotation in the space  \mbox{$\mathbb{R}^5$}, and $V_{\textbf{\textit{R}}} \in G < \SO(4)$ is the 3D rotation representation in the filter bank output space:
\label{eq:steerable_neurons}
\begin{equation}
	\label{eq:V_R}
	V_{\textbf{\textit{R}}} = \textup{\textbf{M}}^\top \textbf{\textit{R}}_O\, \textbf{\textit{R}}\, \textbf{\textit{R}}_O^{\top} \textup{\textbf{M}} ~,
\end{equation}
where $\textup{\textbf{M}} \in \SO(4)$ is a change-of-basis matrix that holds the homogeneous coordinates of the tetrahedron vertices (scaled by $1/2$) in its columns as\vspace{-5pt}
\begin{equation}
	\label{eq:basis_matrix}
	\textbf{M}  = \frac{1}{2}
	\begin{bmatrix}
		1 &  \phantom{-}1 & -1             &  -1   \\
		1 & -1            &  \phantom{-}1  &  -1   \\
	    1 & -1            & -1             &  \phantom{-}1   \\
	    1 &  \phantom{-}1 &  \phantom{-}1  &  \phantom{-}1   \\
	\end{bmatrix} ~.
\end{equation}
We will use the equivariant filter bank output as a replacement for 3D points, \eg, in vector neural networks (VNNs) by Deng \emph{et~al.\@}~\cite{deng2021vector}.

\subsection{Vector neurons}
\label{sec:vector_neurons}
\textit{Vector neurons} (VNs) \cite{deng2021vector} are designed for processing data embedded in $\mathbb{R}^3$ and produce an ordered set of 3D vectors $\textbf{y} \in \mathbb{R}^3$ as output.
Taking a point cloud $\mathcal{X}\in\mathbb{R}^{N\times 3}$ as input, a VN extracts vector-list features $\mathcal{Y} = \left\{\textbf{\textit{Y}}_n\right\}_{n=1}^{N}\in \mathbb{R}^{N \times C \times 3}$, where $\textbf{\textit{Y}} \in \mathbb{R}^{C\times3}$ is a vector-feature and $C$ is the number of latent channels.

Specifically, a linear layer $f_{\text{lin}}(\,\cdot\,;\textbf{W})$ comprised of VNs is defined by means of a weight matrix $\textbf{W} \in \mathbb{R}^{C' \times C}$ acting on a vector-feature $\textbf{\textit{Y}} \in \mathcal{Y}$ as $f_{\text{lin}}(\textbf{\textit{Y}};\textbf{W})=\textbf{W}\textbf{\textit{Y}}$, and is SO(3)-equivariant since
\begin{equation}
    \label{eq:vn_linear}
    f_{\text{lin}}(\textbf{\textit{Y}}\textbf{\textit{R}};\textbf{W}) = \textbf{W}\textbf{\textit{Y}}\textbf{\textit{R}} = f_{\text{lin}}(\textbf{\textit{Y}};\textbf{W})\textbf{\textit{R}} = \textbf{\textit{Y}}'\textbf{\textit{R}},
\end{equation}
where $\textbf{\textit{R}}\in \SO(3)$ and $\textbf{\textit{Y}}' \in \mathbb{R}^{C' \times 3}$.

Deng \emph{et~al.\@}~\cite{deng2021vector} also presented how common neural network operations, such as batch norm \cite{ioffe2015batch}, pooling, and non-linearities, can be adopted for VNs, and how VNs can be used in other point cloud processing networks.
In particular, their VN-DGCNN modifies the permutation-equivariant edge convolution of the predecessor DGCNN \cite{wang2019dynamic} by computing adjacent edge features $\textbf{\textit{E}}'_{nm}\in\mathcal{E}$ of vector-list representations $\textbf{\textit{Y}}_n \in \mathbb{R}^{C \times 3}$, followed by a local SO(3)-equivariant pooling as
\begin{equation}
    \label{eq:vn_edge_conv}
    \textbf{\textit{E}}'_{nm} = l_\textup{VN-nonlin}(\Theta (\textbf{\textit{Y}}_m - \textbf{\textit{Y}}_n) + \Phi \textbf{\textit{Y}}_n),
\end{equation}
\begin{equation}
    \label{eq:vn_local_pool}
    \textbf{\textit{Y}}'_n = l_{\textup{VN-pool}\,{m:(n,m)\in\mathcal{E}}}(\textbf{\textit{E}}'_{nm}),
\end{equation}
where $\Theta$ and $\Phi$ are learnable weight matrices, VN-nonlin and VN-pool are the respective equivariant non-linear and pooling layers (see Section~3 in Deng \emph{et~al.\@}~\cite{deng2021vector} for details). 
Notably, average pooling, being a linear operation, maintains rotation-equivariance and helps to achieve higher performance \cite{deng2021vector}.

To summarize, the important properties of a VNN \cite{deng2021vector} are that 1) it is SO(3)-equivariant and produces RI features at the later layers, and 2) the local interaction between the points is modeled by exploiting edges by means of edge convolutions introduced in DGCNN \cite{wang2019dynamic}.

\begin{figure*}
	\centering
	\includegraphics[width=1\linewidth, trim={0cm 0.5cm 0cm 0.5cm}, clip=true]{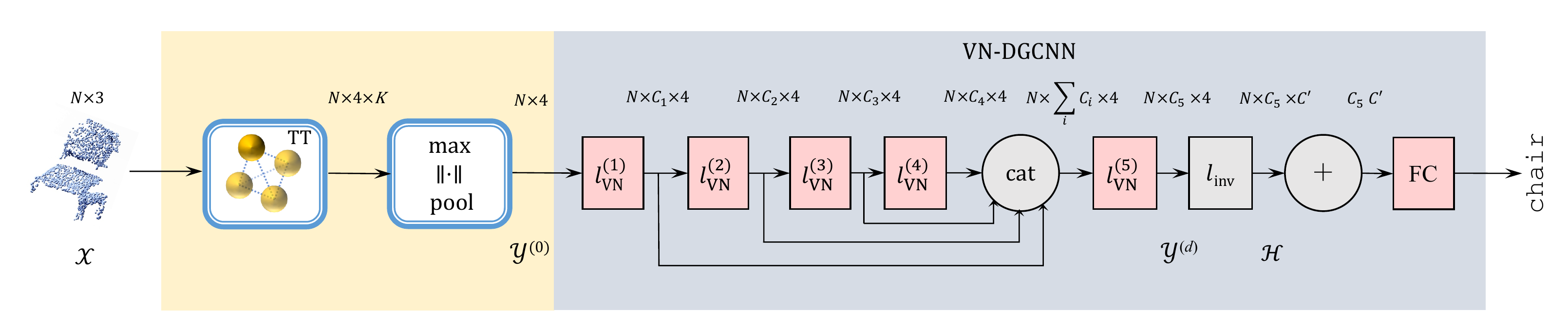}
	\caption{High-level architecture of TetraSphere (for classification): the equivariant TT layer \eqref{eq:tetratransform} is followed by pooling over $K$ steerable spherical neurons and the application of the equivariant VN-DGCNN \cite{deng2021vector}, consisting of $d$ VN-layers $l_\textup{VN}$ \eqref{eq:vn_mlp}, and the block $l_{\textup{inv}}{(\,\cdot\,; \Theta, \Phi)}$ \eqref{eq:tetrasphere_b}, producing invariant features. The first (yellow) block contains the contributions of our work.}
 \vspace{-10pt}
    \label{fig:tetrasphere-b}
\end{figure*}
\section{TetraSphere}
\label{sec:methodology}
\vspace{-5pt}
In this section, we present TetraSphere---a learnable descriptor for O(3)-invariant point cloud processing---based on steerable 3D spherical neurons and the VN-framework (see Figure~\ref{fig:tetrasphere-b}).
Firstly, we note that $\textbf{\textit{R}}$ in \eqref{eq:V_R} can be a reflection, \ie, have a determinant of $-1$, which will change the sign of $\det V_{\textbf{\textit{R}}}$ accordingly, and \eqref{eq:filter_bank_equivariance} will still hold in this case. 
Therefore, $V_{\textbf{\textit{R}}} \in G < \Og(4)$ and steerable neurons \eqref{eq:sphere_filter_bank} are O(3)-equivariant.
The same applies to vector neurons: \eqref{eq:vn_linear} holds even if $\det \textbf{\textit{R}} = -1$, which means that vector neurons are also O(3)-equivariant.

Our overall approach consists of two steps: 1) we extract O(3)-equivariant features, and 2) we obtain O(3)-invariant representations from them.
As the first step, we perform TetraTransform (TT), \ie, lift the 3D input to a specific 4D space spanned by what we call a \textit{tetra-basis} (see Figure~\ref{fig:tetratransform}).
Transforming points in the tetra-basis implies embedding a 3D rotation/reflection into a proper subgroup of O(4), as $V_{\textbf{\textit{R}}} \in G < \Og(4)$.
Since the entire theory of VNs \cite{deng2021vector} applies to $\mathbb{R}^4$ and O(4) exactly the same way it does to $\mathbb{R}^3$ and O(3), we plug our TetraTransform into VNs of dimension $4$ and achieve O(3) invariance.
Note, however, that the VN layers operating on 4D vectors in our model maintain the equivariance under the subgroup $G < \Og(4)$.

\subsection{Learning O(3)-equivariant features}
\label{sec:so3-equiv_features}
\paragraph{TetraTransform} The first layer $l^{(0)}$ is formed by the TT layer $l_{\textup{TT}}(\,\cdot\,;\textbf{S}): \mathbb{R}^{N \times 3} \rightarrow \mathbb{R}^{N \times  4 \times K}$, consists of $K$ steerable spherical neurons $B(\textbf{\textit{S}}_k)$ \eqref{eq:sphere_filter_bank}, representing a $K\times 5$ learnable weight matrix $\textbf{S}$. 

TT first takes in a point cloud of 3D points $\mathcal{X} \in \mathbb{R}^{N \times 3}$ and embeds them in the conformal space $\mathbb{R}^5$ according to  \eqref{hypersphere_in_r}, resulting in $\left\{\textbf{\textit{X}}_n\right\}_{n=1}^{N} \in \mathbb{R}^{N \times 5}$. 

Following the structure of point cloud processing networks \cite{qi2017pointnet, wang2019dynamic}, the subsequent application of the steerable spherical neurons \eqref{eq:sphere_filter_bank} as $B(\textbf{\textit{S}}_k) \,\textbf{\textit{X}}$ is shared across points, thus making the output
\begin{equation}
   \label{eq:tetratransform}
   \mathcal{Y}^{(0)} = l_\text{TT}(\mathcal{X}; \textbf{S}) \in \mathbb{R}^{N \times 4 \times K}
\end{equation}
both rotation- and permutation-equivariant.
Importantly, thanks to the embedding of vectors \eqref{hypersphere_in_r}, $l_{\textup{TT}}(\,\cdot\,;\textbf{S})$ is a non-linear layer, which is essential for neural networks.
\vspace{-12pt}
\paragraph{Tetra-basis projections} Note that each of the $K$ steerable spherical neurons \eqref{eq:sphere_filter_bank} in $l_{\textup{TT}}(\,\cdot\,;\textbf{S})$ has its own representation of a 3D rotation $\textbf{\textit{R}}$, given as $V_{\textbf{\textit{R}}}^k \in G < \Og(4),~~k\in\{1,\dots K\}$, due to the rotation $\textbf{\textit{R}}_O$ in \eqref{eq:V_R} (and \eqref{eq:sphere_filter_bank}) being computed from a learnable $\textbf{\textit{S}}_k$. 
In fact, we see $\mathcal{Y}^{(0)}$ as a collection of $N$ rotation-equivariant 4D vectors in $K$ different tetra-bases.
This must be taken into consideration when transforming $\mathcal{Y}^{(0)}$ so as to preserve equivariance.
\vspace{-12pt}
 \paragraph{Aggregating over tetra-bases} 
The case $K=1$ corresponds to a non-linear change of the coordinate system from 3D to a 4D space spanned by the tetra-basis.
However, to accumulate the features captured in the $K>1$ tetra-bases, we need to consider aggregation operators that respect equivariance.
In our work, we propose to use \textit{maximum} pooling over $K$ steerable neurons/tetra-bases for each of the input $N$ points, thus selecting one of the $K$ 4D outputs of the TT layer, indexed with $k^*$, and define this operation as follows:
\begin{equation}
   \label{eq:max_pooling}
   l_\text{pool}(\mathcal{Y}^{(0)}) = \mathcal{Y}^{(0)}_{:,\, :,\,k^*}, ~~ k^* = \mathop{\mathrm{mode}}_n \, \arg\max_{k}\lVert \mathcal{Y}^{(0)}_{n,\, :,\,k}\rVert.
\end{equation}
For an input point cloud, this operation corresponds to the selection of the $k^*$-th steerable neuron, the output of which has the maximum $l^2$-norm for the majority of points. 
Since each of the $K$ steerable neurons \eqref{eq:sphere_filter_bank} is O(3)-equivariant and hence, preserves the $l^2$-norm of the output, the proposed selection of one of them is O(3)-invariant, thereby respecting the equivariance of the model.
\vspace{-12pt}
\paragraph{Deeper equivariant propagation}
We proceed by adding the O(3)-equivariant VN-framework \cite{deng2021vector}, reviewed in Section~\ref{sec:vector_neurons},  on top of $l_\text{TT}$: we apply VNs to the (pooled over $K$) first layer output $\mathcal{Y}^{(0)}$, which we, therefore, need to view as a list of vector-features $\mathcal{Y}^{(0)} = \left\{\textbf{\textit{Y}}_n\right\}_{n=1}^{N}\in \mathbb{R}^{N \times 4}$. 

We can thus extend VNs \cite{deng2021vector} to operate on our specific 4D vectors, contained in $\mathcal{Y}^{(0)}$. 
Obviously, a linear layer comprised of VNs 
$f_{\text{lin}}(\,\cdot\,;\textbf{W})$ is also equivariant under $V_{\textbf{\textit{R}}} \in G < \Og(4)$.
By replacing $\textbf{\textit{R}}$  in \eqref{eq:vn_linear} with $V_{\textbf{\textit{R}}}$ in \eqref{eq:V_R}, and keeping in mind that vector-features $\textbf{\textit{Y}}$ contain now 4D vectors, we see that \eqref{eq:vn_linear} holds.
The same applies to other equivariant VN-layers (\eg, non-linearities, batch norm); see Deng \emph{et~al.\@}~{\cite{deng2021vector}}.

We denote a consequent application of O(3)-equivariant (and non-linear) edge convolution (EC) \eqref{eq:vn_edge_conv} and pooling \eqref{eq:vn_local_pool} layers as $l_{\textup{VN}}(\,\cdot\,;\Theta,\Phi): \mathbb{R}^{N \times C \times 4} \rightarrow \mathbb{R}^{N \times C' \times 4}$. 
In general, the $d$-th VN-layer taking $\mathcal{Y}^{(d)} \in \mathbb{R}^{N \times C \times 4}$ as input produces an O(3)-equivariant and permutation-equivariant feature map
\begin{equation}
    \label{eq:vn_mlp}
    \mathcal{Y}^{(d+1)} = l_{\textup{VN}}(\mathcal{Y}^{(d)}; \Theta,\Phi) \in \mathbb{R}^{N \times C' \times 4},
\end{equation}
where $C'$ are the latent channels.
Given the (pooled over $K$) TT output \eqref{eq:tetratransform} $\mathcal{Y}^{(0)} \in \mathbb{R}^{N \times 4}$, a VN-layer outputs a feature map $\mathcal{Y}^{(d)} \in \mathbb{R}^{N \times C \times 4}$.
\subsection{O(3)-invariant representations}
\label{sec:o3-invariant_features}
\vspace{-2pt}
The TetraSphere architecture (see Figure~\ref{fig:tetrasphere-b}), presented in this section, performs TT \eqref{eq:tetratransform} as the first step. 

To obtain RI features, we follow related work (\eg, \cite{deng2021vector} and \cite{xu2021sgmnet}) and exploit the fact that the inner product of two roto-equivariant vectors, rotated in $\mathbb{R}^n$ with the same $\textbf{\textit{R}}$, is invariant:
\begin{equation}
    \label{eq:o3_invariance}
     \textbf{\textit{U}} \textbf{\textit{R}} \,(\textbf{\textit{T}} \textbf{\textit{R}})^\top = \textbf{\textit{U}} \textbf{\textit{R}}\textbf{\textit{R}}^\top \textbf{\textit{T}}^\top = \textbf{\textit{U}}\textbf{\textit{T}}^\top = \textbf{H},
\end{equation}
where $\textbf{\textit{U}} \in \mathbb{R}^{C \times n}$, $\textbf{\textit{T}} \in \mathbb{R}^{C'\times n}$, and $\textbf{H} \in \mathbb{R}^{C \times C'}$.
Note that $\textbf{H}$ is O($n$)-invariant since the sign of $\det(\textbf{\textit{R}})$ does not change the equality \eqref{eq:o3_invariance}.
To the best of our knowledge, this has not been observed in prior work.

If we take \eqref{eq:o3_invariance} and consider $\textbf{\textit{U}} \in \mathbb{R}^{C \times 3}$ and $\textbf{\textit{T}} \in \mathbb{R}^{C'\times 3}$ to be 3D vector-features of the same 3D point, but at two different layers with $C$ and $C'$ channels, respectively, we will get the VN-framework approach (see Section~3.5 in Deng~\emph{et~al.\@}~\cite{deng2021vector}). 
In this case, we refer to \eqref{eq:o3_invariance} as a \textit{point-wise} inner product of features.
We adopt this procedure to our 4D vectors:
In the first step, TT \eqref{eq:tetratransform} produces $\mathcal{Y}^{(0)} \in \mathbb{R}^{N \times 4 \times K}$. 
We then apply pooling over $K$ spheres and a desired number of VN-layers \eqref{eq:vn_mlp} to it, obtaining $\mathcal{Y}^{(d)} \in \mathbb{R}^{N \times C \times 4}$.
To produce RI features, we follow Deng~\emph{et~al.\@}~\cite{deng2021vector} and concatenate $\mathcal{Y}^{(d)}$ with its global mean (over $N$),~~ $\overline{\mathcal{Y}}^{(d)} = \frac{1}{N} \sum_{n}{\mathcal{Y}^{(d)}_n} \in \mathbb{R}^{C \times 4}$, and propagate the result through $m$ additional VN-layers to obtain $\mathcal{Y}^{(d+m)} \in \mathbb{R}^{N \times C' \times 4}$.
We then extract matrices $\textbf{\textit{U}} \in \mathbb{R}^{C \times 4}$ from $\mathcal{Y}^{(d)}$ and $\textbf{\textit{T}} \in \mathbb{R}^{C'\times 4}$ from $\mathcal{Y}^{(d+m)}$ and perform \eqref{eq:o3_invariance} for all $N$.
Note that the complexity of this product is linear, \ie, $\mathcal{O}(N)$, as opposed to, \eg, the quadratic complexity of the product in the SGM approach \cite{xu2021sgmnet}.

We denote the propagation from VN-layer $d$ to layer $d+m$ with the subsequent point-wise product as a block $l_{\textup{inv}}{(\,\cdot\,; \Theta, \Phi)}: \mathbb{R}^{N \times C \times 4} \rightarrow \mathbb{R}^{N \times C \times C'}$, where $\Theta$ and $\Phi$ denote the learnable parameters of the VN-layers. 
In practice, we select $C' = 3$ following the original VN-approach \cite{deng2021vector}.

 In the case of a single VN-layer \eqref{eq:vn_mlp} following after the TT-layer \eqref{eq:tetratransform}, we describe TetraSphere operating on $\mathcal{X}\in \mathbb{R}^{N \times 3}$ as\vspace{-10pt}
 \begin{equation}
    \label{eq:tetrasphere_b}
    \mathcal{H} = l_{\textup{inv}}(\,l_{\textup{VN}}(\,{l}_\textup{pool}(\,l_\text{TT}(\mathcal{X};\textbf{S})\,)\,)\,),
\end{equation}
where $\mathcal{H} \in \mathbb{R}^{N \times C \times C'}$ is an O(3)-invariant and permu\-tation-equivariant descriptor of $\mathcal{X}$, that can be used for various point-cloud analysis tasks.
\section{Experiments}
\label{sec:experiments}
\vspace{-5pt}
In this section, we conduct experiments with TetraSphere based on the rotation-equivariant VN-DGCNN architecture \cite{deng2021vector}. 
We evaluate our model using both synthetic and real-world 3D data and compare it with other methods.
\vspace{-2pt}
\subsection{Datasets and tasks}
\begin{figure}
	\centering
	\includegraphics[width=\linewidth, trim={0cm 1.3cm 0cm 0cm}, clip=true]{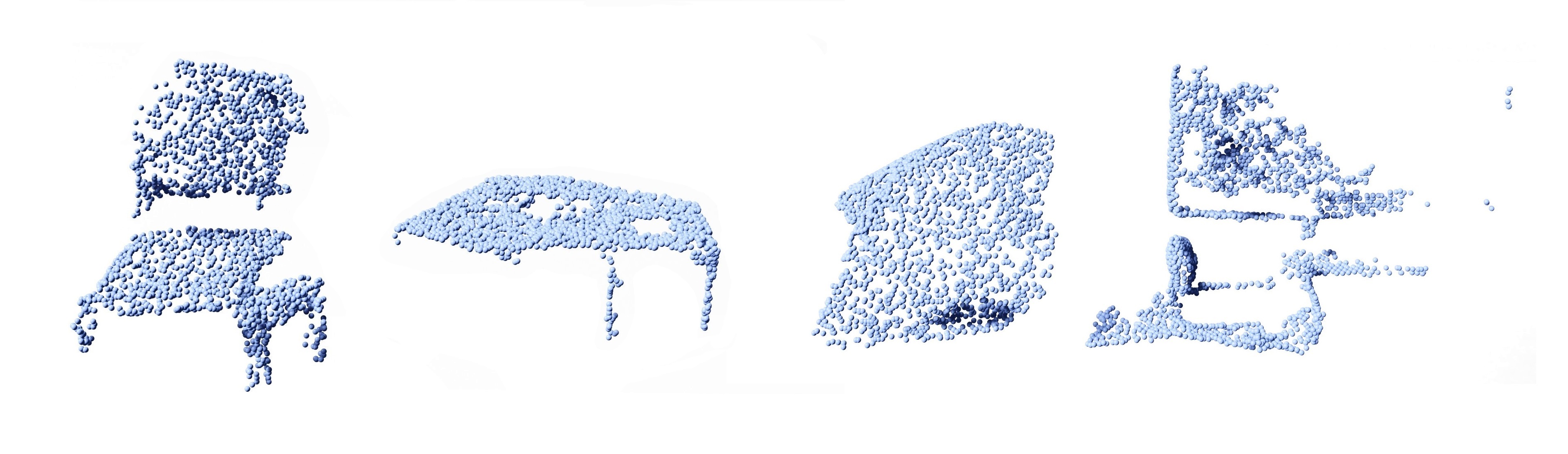}
	\caption{Examples of the objects from the hardest subset of ScanObjectNN~\cite{uy-scanobjectnn-iccv19}: \textit{chair}, \textit{table}, \textit{pillow}, and \textit{display}.}
 \vspace{-15pt}
    \label{fig:sobjnn}
\end{figure}

\vspace{-3pt}
\paragraph{Real data classification} 
We first consider the task of classification and evaluate our method on real-world indoor scenes. 
For this, we use ScanObjectNN \cite{uy-scanobjectnn-iccv19}, which consists of 2902 unique object instances belonging to 15 classes. 
We employ two subsets: the easiest, called \textit{OBJ\_BG} and containing objects with background, and the most challenging subset called \textit{PB\_T50\_RS}, consisting of approximately 15,000 point clouds that undergo 50\% bounding box translation, random rotation around the gravity axis, and random scaling, in which perturbations introduce various levels of partiality to the objects.
We follow the train/test split provided by the original repository\footnote{\url{https://github.com/hkust-vgd/scanobjectnn}}.
Some examples are presented in Figure~\ref{fig:sobjnn}.
We preprocess both datasets the same way, sampling 1024 points per object instance, centering them at the origin, and normalizing them to be within a unit sphere. 
\vspace{-12pt}
\paragraph{Synthetic data classification and part segmentation} In addition, we evaluate our model on the tasks of classifying ModelNet40 data \cite{wu20153d} provided by \cite{qi2017pointnet} that consist of 12,311 CAD models randomly sampled as point clouds comprised of 1024 points, and the task of part segmentation using ShapeNet-part \cite{chang2015shapenet}, consisting 16,881 point clouds of 16 categories partitioned with 50 part labels in total.
We follow \cite{deng2021vector} for the train/test split and sample 2048 points for the model input. 
\vspace{-12pt}
\paragraph{Rotation setup} In general, we employ the following train/test rotation settings, following the general convention \cite{deng2021vector, zhao2019rotation, yu2023rethinking}: $z$/$z$, $z$/$\{\SO(3),~\textup{O}(3)\}$ and $\SO(3)/\SO(3)$, with the second one being the most challenging and the most practical (in which we also include O$(3)$ to test the invariance under the transformations of the full orthogonal group).
Here, $z$ denotes vertical-axis rotation augmentation, SO(3) stands for arbitrary 3D rotations, and O$(3)$ for arbitrary rotations+reflections, all generated and applied to the input shapes during training/testing.
Note that the \textit{PB\_T50\_RS} subset of ScanObjectNN already includes $z$-axis rotation augmentations, and therefore, we do not need to use additional augmentation during training for the $z/\cdot$ scenarios.
\subsection{Architecture and implementation details}
\vspace{-2pt}
We use VN-DGCNN \cite{deng2021vector} as the backbone, with the standard choice of $k=20$ (nearest-neighbor graph computation parameter) for classification and $k=40$ for part-segmentation for all layers and the dropout in the last two fully-connected layers of $0.5$.
We apply VN-LeakyReLU as the learnable equivariant non-linearity in \eqref{eq:vn_edge_conv}, and use average pooling in VN-layers \eqref{eq:vn_local_pool}, given its reported higher performance.
The architecture for part-segmentation experiments is the same, except the VN-DGCNN backbone is adjusted accordingly (as per \cite{deng2021vector}). 
We experiment with different numbers $K$ of steerable spherical neurons in the TT layer and refer to the resulting model simply as \textbf{TetraSphere}.

We adopt the official implementation of Deng~\emph{et~al.\@}~\cite{deng2021vector} to implement our model in PyTorch \cite{paszke2019pytorch}. 
Following Melnyk \emph{et~al.\@}~\cite{melnyk2022steerable}, we initialize the parameters in the TT layer (\ie, the spheres) using the standard initialization for the linear layers in PyTorch.
We use the same hyperparameters for training TetraSphere as the baseline \cite{deng2021vector}: 
We employ SGD with an initial learning rate of $0.1$ and momentum equal to $0.9$, and a cosine annealing strategy for gradually reducing the learning rate to $0.001$, and minimize cross-entropy with smoothed labels.
Like the baseline, we augment the data with random translation in the range $[-0.2, 0.2]$ and scaling in the range $[2/3, 3/2]$ during training.
We train TetraSphere for 1000 epochs for all ScanObjectNN experiments.
Following the baseline, we set the number of epochs to 250 for ModelNet40 classification, and 200 for ShapeNet part segmentation.
The batch size is set to 32.
\vspace{-2pt}
\subsection{Results and discussion}
\label{sec:results_and_discussion}
\vspace{-3pt}
\begin{table}%
    \begin{tblr}{%
            rows={rowsep=0pt,ht=0.9\baselineskip,cmd={\smash},font={\footnotesize}},
           column{3}={blue!5!white},
            column{2,3,4}={c},
            colspec={XQQQ},
            width=\linewidth
        }
        \toprule
        Methods  & $z/z$ & $z/\SO(3)$ & $\SO(3)/\SO(3)$\\
        \midrule
        \SetRow{white}\SetRow{ht=1.1\baselineskip}\SetCell[c=4]{c} Rotation-sensitive\\
        \midrule
        PointCNN \cite{li2018pointcnn}      & 86.1 & 14.6 & 63.7\\
        DGCNN \cite{wang2019dgcnn}          & 82.8 & 17.7 & 71.8\\
        \midrule
        \SetRow{white}\SetRow{ht=1.1\baselineskip}\SetCell[c=4]{c} Rotation-robust\\
        \midrule
        Li \etal \cite{li2021closer}       & 84.3 & 84.3  & 84.3 \\
        PaRINet \cite{chen2022devil}       & 77.8 & 77.8  & 78.1 \\
        PaRINet + PCA \cite{chen2022devil} & 83.3 & 83.3  & 83.3 \\
        Yu \etal \cite{yu2023rethinking}   & -    & \underline{86.6} &  \underline{86.3} \\
        VN-DGCNN \cite{deng2021vector} $^*$ & 83.5 & 83.5 & 84.2 \\
        \textbf{TetraSphere}$_{K=1}$        & 84.7 & 84.7 & 86.2 \\
        \textbf{TetraSphere}$_{K=2}$        & \textbf{87.3} & \textbf{87.3} &	84.9 \\
        \textbf{TetraSphere}$_{K=4}$        & 84.5 & 84.5 & \textbf{87.1} \\
        \textbf{TetraSphere}$_{K=8}$        & 86.2 & 86.2 &	84.9 \\
        \textbf{TetraSphere}$_{K=16}$       & 85.4 & 85.4 & 85.9 \\
        \bottomrule
    \end{tblr}
    \caption{Classification acc. (\%) on the real-world objects from the \textit{OBJ\_BG} (easiest) subset of ScanObjectNN under different train/test settings of rotation augmentation. The overall best results are presented in \textbf{bold}, and the second-best are \underline{underlined}. We evaluated methods marked with $^*$ using their open-source implementation.}
    \label{tab:object_bg}
\end{table}
\begin{table}%
    \begin{tblr}{%
            rows={rowsep=0pt,ht=0.9\baselineskip,cmd={\smash},font={\footnotesize}},
           column{3}={blue!5!white},
           column{2,3,4}={c},
           colspec={XQQQ},
           width=\linewidth
        }
        \toprule
        Methods  & $z/z$ & $z/\SO(3)$ & $\SO(3)/\SO(3)$\\
        \midrule
        \SetRow{white}\SetRow{ht=1.1\baselineskip}\SetCell[c=4]{c} Rotation-sensitive\\
        \midrule
        PointCNN \cite{li2018pointcnn}      & 78.5 & 14.9 & 51.8\\
        DGCNN \cite{wang2019dgcnn}          & 78.1 & 16.1 & 63.4\\
        \midrule
        \SetRow{white}\SetRow{ht=1.1\baselineskip}\SetCell[c=4]{c} Rotation-robust\\
        \midrule
        3D-GFE \cite{chou20213dgfe}         & 73.5 &72.7&73.5\\
        Li \etal \cite{li2021closer}  $^*$      & 74.6 & 74.6  & 74.9 \\
        PaRINet \cite{chen2022devil}  $^*$      & 71.6 &  71.6  &  72.2   \\
        Yu \etal \cite{yu2023rethinking} $^*$    & 77.2 & 77.2 & 77.4   \\
        VN-DGCNN \cite{deng2021vector} $^*$      & 77.9 & 77.9 & 78.5 \\
        \textbf{TetraSphere}$_{K=1}$  & 78.5 & 78.5 & \underline{78.7} \\ 
        \textbf{TetraSphere}$_{K=2}$  & \underline{78.9} & \underline{78.9} & \textbf{79.0} \\ 
        \textbf{TetraSphere}$_{K=4}$  & \textbf{79.2} & \textbf{79.2}	& \textbf{79.0} \\ 
        \textbf{TetraSphere}$_{K=8}$  & 78.7 & 78.7	& \textbf{79.0} \\ 
        \textbf{TetraSphere}$_{K=16}$ & 78.8 & 78.8	& \textbf{79.0} \\ 
        \bottomrule
    \end{tblr}
    \caption{Classification acc. (\%) on the real-world objects from the \textit{PB\_T50\_RS} (hardest) subset of ScanObjectNN under different train/test settings of rotation augmentation. The overall best results are presented in \textbf{bold}, and the second-best are \underline{underlined}. We evaluated methods marked with $^*$ using their open-source code.}
    \vspace{-12pt}
    \label{tab:scanobjectnn}
\end{table}
\begin{table}
\vspace{-5pt}
\centering
    \begin{tblr}{%
        rows={rowsep=0pt,ht=0.9\baselineskip,cmd={\smash},font={\footnotesize}}, 
            column{4}={blue!5!white}, 
            column{7}={blue!5!white},
            column{2,3,4,5,6,7}={c},
            colspec={QQQQQQ},
            width=\linewidth
        }
        \toprule
        & &  \SetCell[c=2]{c} ModelNet40 &   & & \SetCell[c=2]{c} ShapeNet \\
        \cmidrule{3-4}         \cmidrule{6-7} 
        Methods & & $z/z$  & $z/\SO(3)$ & & $z/z$  & $z/\SO(3)$ \\
        \midrule
        TFN \cite{poulenard2021functional} & &  89.7 & 89.7 & & -    & 78.1 \\
        VN-DGCNN \cite{deng2021vector}     & &  89.5 & 89.5 & & 81.4 & 81.4 \\
        \textbf{TetraSphere}$_{K=1}$       & &  89.5 & 89.5 & & 82.1 & 82.1 \\ 
        \textbf{TetraSphere}$_{K=2}$       & &  89.7 & 89.7 & & \textbf{82.3}&  \textbf{82.3} \\
        \textbf{TetraSphere}$_{K=4}$       & &  \underline{90.0} & \underline{90.0} & & \underline{82.2} & \underline{82.2} \\ 
        \textbf{TetraSphere}$_{K=8}$       & &  \textbf{90.5} & \textbf{90.5} & & \textbf{82.3} & \textbf{82.3} \\ 
        \textbf{TetraSphere}$_{K=16}$      & &  89.8 & 89.8 & & \textbf{82.3} & \textbf{82.3} \\ 
        \bottomrule
    \end{tblr}
    \caption{Comparison of rotation-equivariant methods using synthetic noiseless data. Left: Classification accuracy (\%) on the ModelNet40 shapes. Right: Part segmentation of the ShapeNet shapes, mIoU (\%). The best results are presented in \textbf{bold}, and the second-best are \underline{underlined}.}
    \vspace{-15pt}
    \label{tab:shapenet}
\end{table}
The main results of our experiments are presented in Tables~\ref{tab:object_bg},~\ref{tab:scanobjectnn},~and~\ref{tab:shapenet}, where for a fair comparison, we list methods that only use point clouds as input, and no additional information, such as normals or features, or test-time augmentation.
From Table~\ref{tab:object_bg}, in the task of classifying the easier subset of real-world object scans with background (and no perturbations), our method outperforms the baseline VN-DGCNN, especially in the more practical $z/\SO(3)$ scenario, and the recent method by Yu~\etal~\cite{yu2023rethinking},  thus setting a new state-of-the-art performance.
Here, we observe that increasing the number of steerable neurons beyond $K = 2$ does not systematically improve the performance.
In general, the results under the $\SO(3)/\SO(3)$ protocol indicate that additional rotation augmentation when classifying non-perturbed shapes is not required for our method.

The previous best result published in the literature in our comparison for the classification of the perturbed real object scans from the most challenging subset of ScanObjectNN (see Table~\ref{tab:scanobjectnn}) is 3D-GFE \cite{chou20213dgfe}. 
We used the open-source implementations and evaluated the recent related methods, showing that VN-DGCNN exhibits better robustness to perturbations (comparing with the results of the perturbation-free \textit{OBJ\_BG} test in Table~\ref{tab:object_bg}) --- a property we attribute to the equivariant feature extraction scheme of the VN framework.
With our TetraSphere (built upon VN-DGCNN), the performance is further boosted thanks to the 4D representation learning enabled by the TetraTransform layer with $K \geq 1$: TetraSphere achieves state-of-the-art classification performance.

As shown in Table~\ref{tab:shapenet}, TetraSphere outperforms equivariant baselines at the tasks of classifying and segmenting parts of the synthetic shapes.
Our model is only surpassed by PaRINet~\cite{chen2022devil} (the complete tables are presented in the Supplementary Material) and by Yu~\etal~\cite{yu2023rethinking} (only on ModelNet40), both of which TetraSphere exceeds the performance of on the other two real-data benchmarks (see Tables~\ref{tab:object_bg}~and~\ref{tab:scanobjectnn}), even when PaRINet is aided by PCA.
Compared to the proposed equivariant method, the previous state-of-the-art methods degrade when the effects of real data occur: noise, occlusion, and outliers.

We also experimentally verify that TetraSphere is O$(3)$-invariant by applying random reflections in addition to rotations during inference, as shown in Table~\ref{tab:o3_inv}: the accuracies of the TetraSphere evaluated on the data augmented with $z$-axis rotations and O$(3)$-transformations are identical.

Furthermore, we perform an ablation study testing the importance of the 4D representation learned by TetraSpheres as opposed to the baseline VN-DGCNN operating on the original 3D point coordinates appended with a fourth, invariant, component. 
For this, we append each input point, $\textbf{x} \in \mathbb{R}^3$ with its norm, thus making the input 4D, \ie, $[\textbf{x}, \lVert \textbf{x} \rVert]$.
We also compare our model to VN-DGCNN$_{+l_0}$ and VN-DGCNN$_{+l_0}$(${[\textbf{x}, \lVert \textbf{x} \rVert]}$) models, in which the baseline VN-DGCNN has an additional 0-th (equivariant) VN-layer inserted at the beginning, which makes the model have the same depth as TetraSphere and adds $366$ 
parameters to the baseline ($~0.01265\%$).
As presented in Table~\ref{tab:o3_inv}, TetraSphere outperforms the baseline.  
\vspace{-9pt}
\paragraph{Learned Tetra-selection}
\vspace{-5pt}
Even though our proposed equivariant pooling~\eqref{eq:max_pooling} allows for selecting different steerable neurons (from the $K$ available ones) for different inputs, we found that TetraSphere learns to select the same neuron (\ie, tetra-basis) for \textit{all} inputs.
As we present in the Supplementary Material, our model does so by learning all but one $\gamma$ parameter of the spherical decision surfaces (see~\eqref{eq:spherical_neuron}) defining the steerable neuron~\eqref{eq:sphere_filter_bank} in the TT layer \eqref{eq:tetratransform}, to be close to 0.
This renders the $l^2$-norms of the 4D activations of the corresponding steerable neurons negligible, thus making TetraSphere always select the steerable neuron with a non-zero $\gamma$.
This means that one can prune the network after training, based on the learned parameters of the TT layer, effectively obtaining $K=1$ at inference, thus reducing the computational time.
\vspace{-10pt}
\paragraph{Complexity analysis}
\vspace{-5pt}
Since TetraSphere is predominantly based on VN-DGCNN, which is in turn based on DGCNN, its computational complexity is not fundamentally different from other methods \cite{deng2021vector}. 
The parameter difference between TetraSphere and the baseline VN-DGCNN is negligible: the former has only one additional TetraTransform layer (see Figure~\ref{fig:tetrasphere-b}), containing $K$ learnable spheres with $5$ parameters each (less than $0.0002\%$ of the baseline size). 
The time complexity difference between the two comes from the usage of 4D vectors by TetraSphere and partially from the TetraTransform operations---the application of the steerable neurons \eqref{eq:sphere_filter_bank}.
Benchmarked on NVIDIA A100, a forward pass through VN-DGCNN takes 5.1ms vs. 6.6ms ($\Delta=\,$1.5ms) through VN-DGCNN operating on 4D vectors (${[\textbf{x}, \lVert \textbf{x} \rVert]}$) vs. 7.9ms ($\Delta=\,$1.3ms) through our implementation of TetraSphere.
\begin{table}
 \begin{tblr}{%
        rows={rowsep=0pt,ht=0.9\baselineskip,cmd={\smash},font={\footnotesize}},          
            column{2,3}={c},
            colspec={XQQ},
            width=\linewidth
        }
        \toprule
        Methods & $z/z$ & $z/\textup{O}(3)$\\
        \midrule
        VN-DGCNN & $82.8 \pm 0.6 ~(83.5)$ & $82.8 \pm 0.6 ~(83.5)$ \\
        VN-DGCNN(${[\textbf{x}, \lVert \textbf{x} \rVert]}$)  & $82.7 \pm 1.6 ~(84.5)$ & $82.7 \pm 1.6 ~(84.5)$ \\
        VN-DGCNN$_{+l_0}$ & $83.3 \pm 0.8 ~(83.8)$ &  $83.3 \pm 0.8 ~(83.8)$ \\
        VN-DGCNN$_{+l_0}$(${[\textbf{x}, \lVert \textbf{x} \rVert]}$) &  $82.7 \pm 0.1 ~(82.8)$ & $82.7 \pm 0.1 ~(82.8)$ \\
        \textbf{TetraSphere}$_{K=2}$  & $\textbf{85.5} \pm \textbf{2.2} ~(\textbf{87.3})$ & $\textbf{85.5} \pm \textbf{2.2} ~(\textbf{87.3})$ \\
        \bottomrule
    \end{tblr}
    \caption{O(3)-test and ablation: Classification acc. (mean and std over 3 runs with the best result in parentheses, \%) on ScanObjectNN \textit{OBJ\_BG} objects under different train/test transformation settings.}
    \vspace{-10pt}
    \label{tab:o3_inv}
\end{table}
\vspace{-7pt}
\section{Conclusion}
\label{sec:discussion}
\vspace{-5pt}
In this paper, we proposed the $\Og(3)$-invariant TetraSphere descriptor as an embedding of steerable 3D spherical neurons into 4D vector neurons.
To the best of our knowledge, we use the steerable neurons in an \textit{end-to-end} approach for the first time, thereby unveiling their practical utility.
TetraSphere sets a new state-of-the-art performance on the task of classifying randomly rotated 3D objects from the challenging real-world ScanObjectNN dataset, and the best results among equivariant methods for classifying and segmenting parts of randomly rotated synthetic shapes from ModelNet40 and ShapeNet, respectively.
We look forward to our work paving the path to geometrically justified and more robust handling of real-world 3D data.
\vspace{-15pt}

\paragraph{Acknowledgments}
{\small This work was supported by the Wallenberg AI, Autonomous Systems and Software Program (WASP), by the Swedish Research Council through a grant for the project Uncertainty-Aware Transformers for Regression Tasks in Computer Vision (2022-04266), and the strategic research environment ELLIIT. The computations were enabled by resources provided by the National Academic Infrastructure for Supercomputing in Sweden (NAISS) partially funded by the Swedish Research Council through grant agreement no. 2022-06725.3, and by the Berzelius resource provided by the Knut and Alice Wallenberg Foundation at the National Supercomputer Centre.}

{
    \small
    \bibliographystyle{ieeenat_fullname}
    \bibliography{main}
}

\clearpage
\setcounter{page}{1}
\maketitlesupplementary
\begin{figure}[t]
	\centering
	\includegraphics[width=1\linewidth]{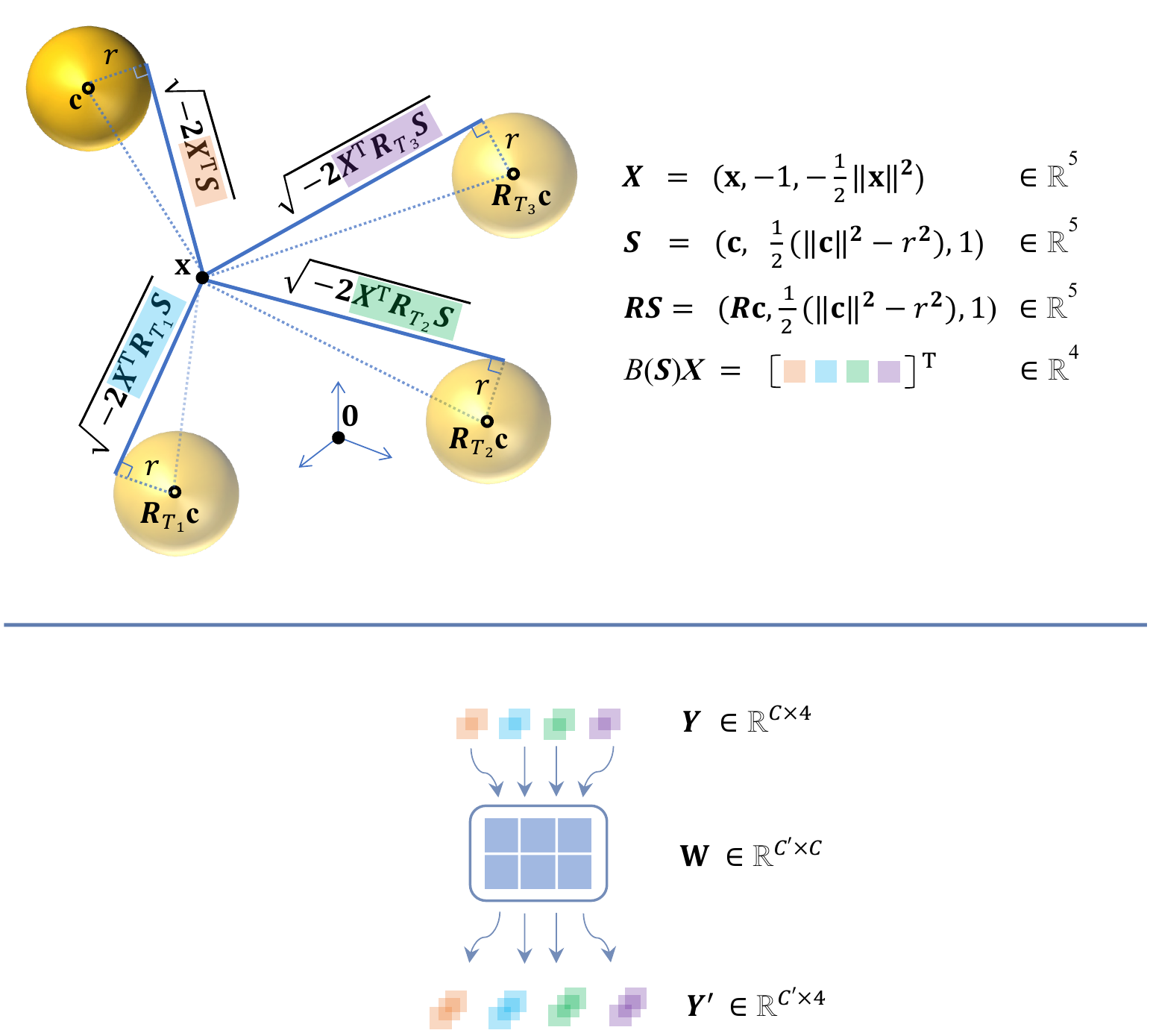}
	\caption{(Best viewed in color.) \textbf{Top}: Tetra-basis projection is the output of a steerable 3D spherical neuron~\citep{melnyk2022steerable}. Without loss of generality, consider one ($K=1$) steerable spherical neuron $B(\textbf{\textit{S}})$ (see Section~\ref{sec:steerable_3d_neurons}) with $\textbf{\textit{R}}_{O} = \textbf{I}_5$, and the input point $\textbf{x}$ that happens to lie outside of the sphere $(\textbf{c}, r)$ with the learnable parameter vector $\textbf{\textit{S}}$ (assume $\gamma=1$, and thus $\Tilde{\textbf{\textit{S}}}=\textbf{\textit{S}}$; see Section~\ref{sec:spherical_neurons}) and its three rotated copies. Then the projection of $\textbf{x}$ in the tetra-basis $B(\textbf{\textit{S}})$ is the vector $B(\textbf{\textit{S}})\textbf{\textit{X}}$ consisting of four scalar activations $\textbf{\textit{X}}^\top \textbf{\textit{R}}_{T_i} \textbf{\textit{S}}$ of the respective spherical decision surfaces. Each activation determines the respective cathetus length, as per~\cite{melnyk2020embed}. \textbf{Bottom}: Vector neurons~\citep{deng2021vector} preserve the spatial dimension ($4$ in our case) and alter the latent dimension $C$ of the feature $\textbf{\textit{Y}}$, see \eqref{eq:vn_linear}.}
    \label{fig:tetra-basis projection}
\end{figure}

\section{Additional illustrations}
\label{sec:additional_illustrations}
In order to help the reader to understand the main concepts of our approach, \ie, prior work (steerable) spherical neurons~\citep{melnyk2022steerable} and vector neurons~\citep{deng2021vector}, as well as 4D tetra-basis projections (see Figure~\ref{fig:tetratransform} and Section~\ref{sec:so3-equiv_features}), we provide illustrations in Figure~\ref{fig:tetra-basis projection}.

\section{Learned Tetra-selection}
\label{sec:learned_parameters}
\begin{figure}
	\centering
	\includegraphics[width=\linewidth]{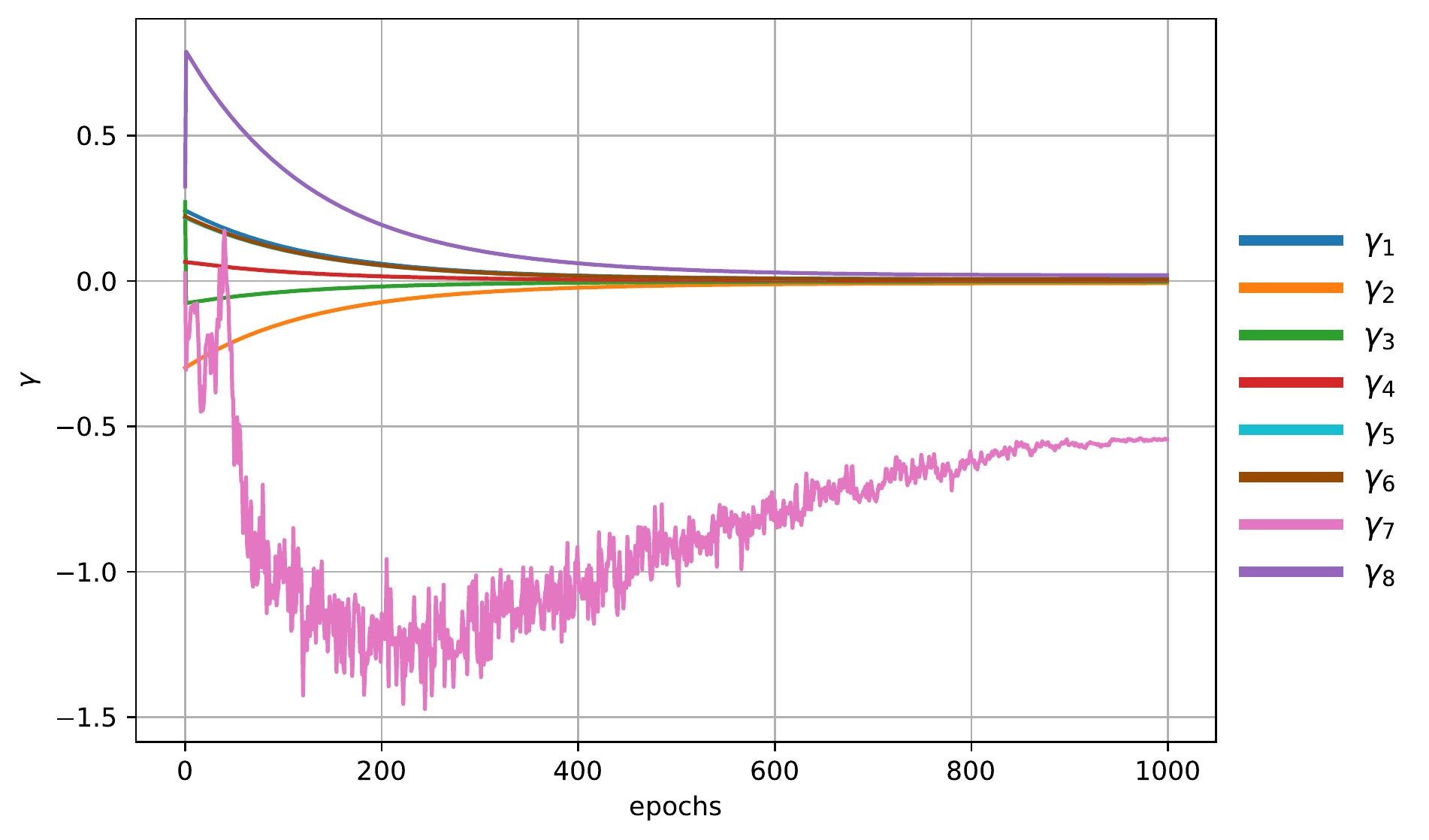}
	\caption{Learned $\gamma$ parameters for \textbf{TetraSphere}$_{K=8}$ trained on the \textit{OBJ\_BG} subset of ScanObjectNN (see Table~\ref{tab:object_bg}). All but $\gamma_{7}$ converge close to 0.}
    \label{fig:gamma_objbg}
\end{figure}

\begin{figure}
	\centering
	\includegraphics[width=\linewidth]{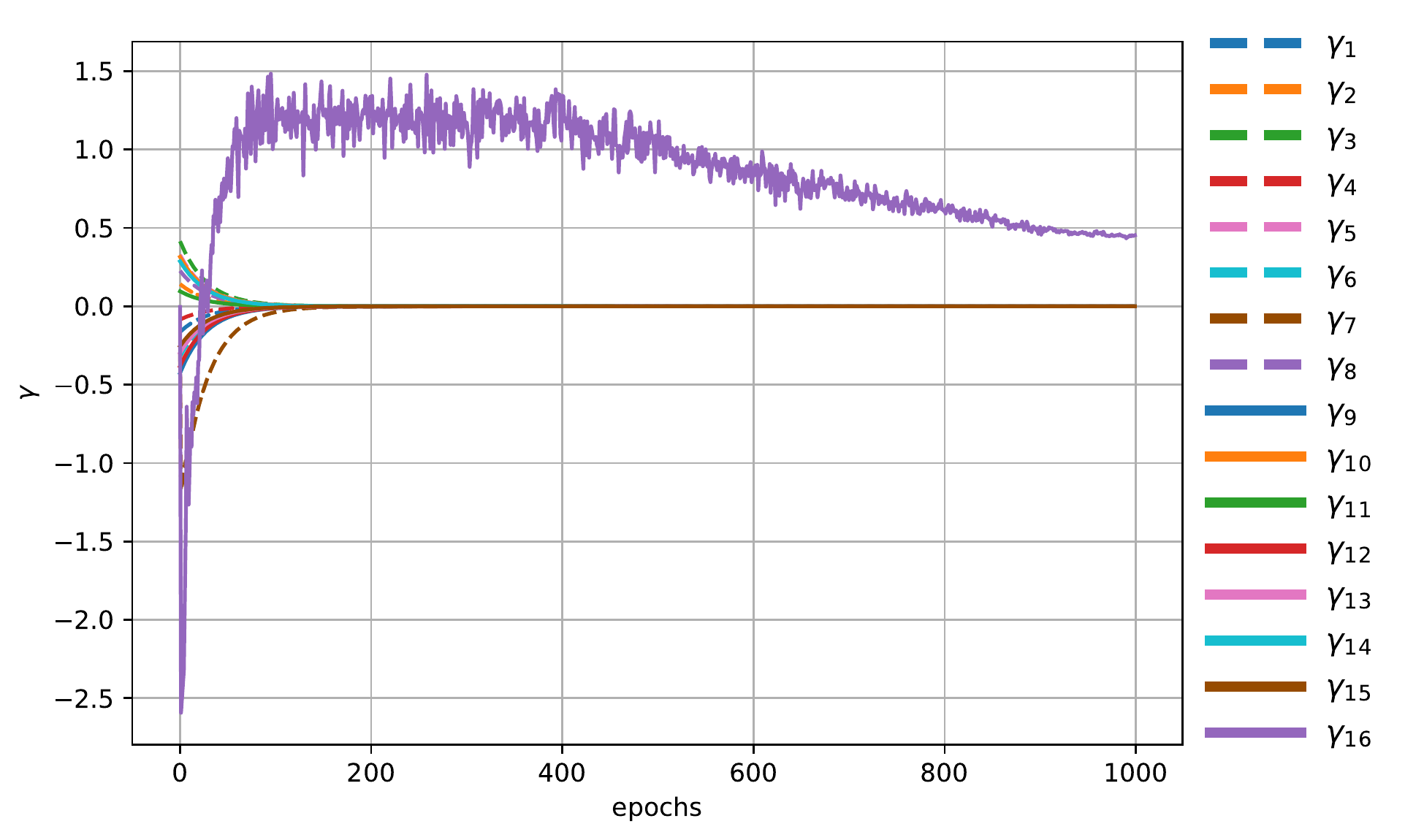} 
	\caption{Learned $\gamma$ parameters for \textbf{TetraSphere}$_{K=16}$ trained on the \textit{PB\_T50\_RS} (see Table~\ref{tab:scanobjectnn}) of ScanObjectNN. All but $\gamma_{16}$ converge close to 0.}
    \label{fig:gamma_pb}
\end{figure}

\vspace{-5pt}
In this section, we present the Tetra-selection discussed in Section~\ref{sec:results_and_discussion}.
As we can see from Figures~\ref{fig:gamma_objbg}~and~\ref{fig:gamma_pb}, \textbf{TetraSphere} learns all but one $\gamma$ parameter of the spherical decision surface (see~\eqref{eq:spherical_neuron}), defining the steerable neuron~\eqref{eq:sphere_filter_bank}, to be close to 0, effectively always selecting one tetra-basis (out of $K$) during inference.
We attribute the increased performance for $K > 1$ (see Tables~\ref{tab:object_bg},~\ref{tab:scanobjectnn}, and \ref{tab:shapenet}), to the higher chance of selecting a better initialization of the steerable neuron parameters.

\section{Synthetic data results}
\label{sec:results_supp}
We present a complete comparison of the methods trained on synthetic data to perform classification and part segmentation in Tables~\ref{tab:modelnet40_supp} and \ref{tab:shapenet_supp}, respectively.
Our \textbf{TetraSphere} achieves the best performance among equivariant methods in both tasks, consistently outperforming VN-DGCNN. 

Only the two RI methods PaRINet~\cite{chen2022devil} and Yu~\etal~\cite{yu2023rethinking} outperform tetrasphere in the former case and only PaRINet in the latter. 
Note that TetraSphere outperforms both PaRINet and Yu~\etal on the other two real-data benchmarks (see Tables~\ref{tab:object_bg}~and~\ref{tab:scanobjectnn}).
\pagebreak

\clearpage
\begin{table}%
    \begin{tblr}{%
            rows={rowsep=0pt,ht=0.9\baselineskip,cmd={\smash},font={\footnotesize}},
            column{3}={blue!5!white},
            column{2,3,4}={c},
            colspec={X >{\hspace{7pt}} QQQ},
            width=\linewidth
        }
        \toprule
        Methods  & $z/z$ & $z/\SO(3)$ & $\SO(3)/\SO(3)$\\
        \midrule
        \SetRow{ht=1.1\baselineskip}\SetCell[c=4]{c} Rotation-sensitive \\
        \midrule
        PointCNN   \cite{li2018pointcnn}                              & \textbf{92.5} & 41.2 & 84.5              \\
        DGCNN      \cite{wang2019dgcnn}                               & 90.3 & 33.8 & 88.6              \\
        \midrule
        \SetRow{ht=1.1\baselineskip}\SetCell[c=4]{c} Rotation-invariant \\
        \midrule
        3D-GFE \cite{chou20213dgfe}                                   & 88.6 & 89.4 & 89.0    \\
        Li \textit{et al.} \cite{li2021closer}                        & 90.2 & 90.2 & 90.2    \\
        Yu \etal \cite{yu2023rethinking}                              & 91.0 & \underline{91.0} & \underline{91.0} \\
        PaRINet \cite{chen2022devil}                                  & \underline{91.4} & \textbf{91.4} & \textbf{91.4}    \\
        \midrule
        \SetRow{ht=1.1\baselineskip}\SetCell[c=4]{c} Rotation-equivariant \\
        \midrule
        TFN \cite{poulenard2021functional}                            & 89.7 & 89.7 & 89.7 \\
        VN-DGCNN \cite{deng2021vector}                                & 89.5 & 89.5 & 90.2    \\ 
        \textbf{TetraSphere}$_{K=1}$                                  & 89.5 & 89.5  & 89.9    \\
        \textbf{TetraSphere}$_{K=2}$                                  & 89.7 & 89.7 &  90.0   \\
        \textbf{TetraSphere}$_{K=4}$                                  & 90.0 & 90.0 &  89.5   \\
        \textbf{TetraSphere}$_{K=8}$                                  & 90.5 & 90.5 &  90.3   \\
        \textbf{TetraSphere}$_{K=16}$                                 & 89.8 & 89.8 &  90.0   \\
        \bottomrule
    \end{tblr}
    \vspace{-10pt}
    \caption{Classification acc. (\%) on the ModelNet40 shapes under different train/test settings of rotation augmentation. The overall best results are presented in \textbf{bold}, and the second best are \underline{underlined}. Our \textbf{TetraSphere} sets a new state-of-the-art performance for equivariant baselines.}
    \label{tab:modelnet40_supp}
    \vspace{-2pt}
\end{table}

{\phantom{\lipsum[1]}}
\pagebreak

\begin{table}[t]
\centering
    \begin{tblr}{%
        rows={rowsep=0pt,ht=0.9\baselineskip,cmd={\smash},font={\footnotesize}},            column{3}={blue!5!white},
            column{2,3,4}={c},
            colspec={XQQQ},
            width=\linewidth
        }
        \toprule
        Methods &  $z/z$  & $z/\SO(3)$ &  $\SO(3)/\SO(3)$\\
        \midrule
        \SetRow{white}\SetRow{ht=1.1\baselineskip}\SetCell[c=4]{c} Rotation-sensitive\\
        \midrule
        PointCNN \cite{li2018pointcnn}      & \textbf{84.6} & 34.7 & 71.4 \\
        DGCNN \cite{wang2019dgcnn}          & 82.3 & 37.4 & 73.3\\
        \midrule
         \SetRow{white}\SetRow{ht=1.1\baselineskip}\SetCell[c=4]{c} Rotation-invariant\\
        \midrule
        3D-GFE \cite{chou20213dgfe}  & - & 78.2 & 77.7    \\
        Li \etal \cite{li2021closer} & 81.7 & 81.7 & 81.7 \\
        PaRINet \cite{chen2022devil} & \underline{83.8} & \textbf{83.8} & \textbf{83.8} \\
        Yu~\etal \cite{yu2023rethinking} & - & 80.3 & 80.4 \\
        \midrule
        \SetRow{white}\SetRow{ht=1.1\baselineskip}\SetCell[c=4]{c} Rotation-equivariant\\
        \midrule
        TFN \cite{poulenard2021functional} & - & 78.1 & 78.2 \\
        VN-DGCNN \cite{deng2021vector}  & 81.4 & 81.4 & 81.4 \\
        \textbf{TetraSphere}$_{K=1}$  & 82.1 & 82.1 & 82.3 \\ 
        \textbf{TetraSphere}$_{K=2}$  & 82.3&  82.3 & 82.5 \\ 
        \textbf{TetraSphere}$_{K=4}$  & 82.2 & 82.2 & 82.2 \\ 
        \textbf{TetraSphere}$_{K=8}$  & 82.3 & 82.3 & 82.4 \\ 
        \textbf{TetraSphere}$_{K=16}$ & 82.3 & 82.3 & 82.3 \\ 
        \bottomrule
    \end{tblr}
    \vspace{-5pt}
    \caption{Part segmentation: ShapeNet mIoU (\%). The overall best results are presented in \textbf{bold}, and the second best are \underline{underlined}. Our \textbf{TetraSphere} sets a new state-of-the-art performance for equivariant baselines.}
    \vspace{-15pt}
    \label{tab:shapenet_supp}
\end{table}

{\phantom{\lipsum[1]}}

\end{document}